# Development Of An Android Application For Object Detection Based On Color, Shape, Or Local Features

Lamiaa A. Elrefaei[1,2], Mona Omar Al-musawa[2] and Norah Abdullah Al-gohany[2]

[1]Electrical Engineering Department, Faculty of Engineering at Shoubra, Benha University, Cairo, Egypt
[2]Computer Science Department, Faculty of Computing and Information Technology, King Abdulaziz University, Jeddah, Saudi Arabia

## ABSTRACT

*Object detection and recognition is an important task in many computer vision applications. In this paper an Android application was developed using Eclipse IDE and OpenCV3 Library. This application is able to detect objects in an image that is loaded from the mobile gallery, based on its color, shape, or local features. The image is processed in the HSV color domain for better color detection. Circular shapes are detected using Circular Hough Transform and other shapes are detected using Douglas-Peucker algorithm. BRISK (binary robust invariant scalable keypoints) local features were applied in the developed Android application for matching an object image in another scene image. The steps of the proposed detection algorithms are described, and the interfaces of the application are illustrated. The application is ported and tested on Galaxy S3, S6, and Note1 Smartphones. Based on the experimental results, the application is capable of detecting eleven different colors, detecting two dimensional geometrical shapes including circles, rectangles, triangles, and squares, and correctly match local features of object and scene images for different conditions. The application could be used as a standalone application, or as a part of another application such as Robot systems, traffic systems, e-learning applications, information retrieval and many others.*

## 1. INTRODUCTION

Object detection is one of the important and challenging tasks in computer vision. Sometimes we need to detect parts of an image that have specific features such as color, shape or local features. The goal of this paper is to develop an Android application that is able to detect objects in an image based on its color, shape, or local features. Most of color detection methods process the image in the HSV color model [1]. In which the image is scanned to find a specific color pixels depending on the HSV range of the color. Shape is one of the main sources of information that can be used for object detection. Some applications may require finding only circles or rectangles [2]. Hough Transform and contours are used in the developed application to find a specific shape region. The developed application is detecting circles, rectangles, triangles, and squares. Any shape other than these shapes is detected as "Other". One of the best methods to detect and recognize specific object in a scene image is by studying local features for both images. The BRISK local features is used in the developed application to match an object in another scene. The rest of this paper is organized as follows: Section 2 presents the related Work, Section 3 introduces the used algorithms for object detection based on color, shape, and local features. The experimental results are introduced in section 4. Finally, Section 5 outlines the conclusion of this work and future work.

## 2. RELATED WORK

There are many earlier works that detect objects based only on color features. The authors in [1] applied the color image segmentation technique using thresholding, to detect the flowers from the scene. They relied on the flower color as it is considered a fundamental property of agriculture products. They operated in the HSV color space and by histogram analysis, a flower color is defined. Then, flowers were separated from the background and detected in images by an image segmentation process.

Other researchers detect objects based on shape features. The researchers in [2] recognized two dimensional shapes and also the shapes type in an image. All known shapes are recognized by segmenting images into regions corresponding to individual objects and then a shape factor is determined and used to recognize the shape type. The work presented in [3] detected circle, square and triangle objects in the image. Their method utilizes intensity value from the input image then a binary image is obtained by thresholding using Otsu's method. The noise is removed by median filter and edges are detected using Sobel operator. The unnecessary edge pixels are removed by thinning method. Their method archived 85% accuracy when tested on the chosen database.

Many other researchers detected objects based on both color and shape features. Authors in [4] identified basic geometric shapes and primary colors in a two dimensional image by applying image processing techniques with the assistance of MATLAB. The essential shapes are square, circle, triangle and parallelogram. The process involves conversion of the RGB image to a gray scale image then to a black and white image. The space of the region of the minimum bounding parallelogram is calculated regardless of the angle of rotation of the object and quantitative relation of this space to the area of the object is calculated and compared to the predefined quantitative relation to work out the form of the given object. The dominant color pixels contribution helps to work out the color of the object. The research work in [5] discussed an approach relating digital image processing and geometric logic for the detection of two dimensional shapes of items such as square, circle, rectangle and triangle, also the color of the object. The steps consisting of RGB image to black and white image conversion, color pixel classification for object-background separation, region based filtering and using of a bounding box and its properties for calculative object metrics. The item metrics are compared with preset values that are characteristics of a selected object's shape. The detection of the shape of the objects is created invariant to their revolution. The colors of the objects are recognized by analyzing RGB information of all pixels within each object. Their algorithm was developed and simulated on MATLAB.

One of the best methods to detect and recognize a specific object on a scene image is by studying local features for both images. Distinctive key points that are reliably localized under varying image transformations, such as translation, rotation, scaling and multi viewpoints [6] can be extracted and used for finding the similar keypoints between images, then select the interested object region. For object detection based on local features, many researchers used local features like SURF (Speeded-Up Robust Features), SIFT (Scale Invariant Feature Transform), and BRISK (binary robust invariant scalable keypoints). The authors in [7] proposed object recognition in which SURF algorithm is used for recognizing multiple objects. They measured the object recognition accuracy under variable conditions of image transformations. The research work in [8] introduced the SIFT detector-SURF-descriptor method that is better than SIFT-detector-SIFT-descriptor (classical SIFT) and SURF detector-SURF-descriptor (classical SURF), but gave less accuracy than the SURF–detector-SIFT descriptor method and suffers in its speed when compared to classical SIFT and SURF. The authors in [9] introduced an object recognition system which resolve the rotations of the object, scale changes, and illumination with the help of "SIFT algorithm". Their implementation consists of many phases such as scale space extreme detection, and key point localization. The authors in [10] presented a High-Speed Recognition Algorithm Based on BRISK for Aerial Images. Their experimental results proved the validity of BRISK features under the conditions of rotation, scale, illumination, and viewpoint changes.

In this paper an Android application that is able to detect objects in an image that is loaded from mobile gallery, based on its color, shape, or BRISK local features is developed.

## 3. OBJECT DETECTION ALGORITHMS

This section shows the algorithms used to detect the object depending on its specific features: color features, shape features, or local features.

### 3.1. OBJECT DETECTION BASED ON ITS COLOR

One of the basic methods to detect objects is by analyzing its color. The proposed algorithm for object detection based on its color consists of the following steps:

**Step1:** The RGB color image that is loaded from the mobile gallery is converted to HSV color image.

**Step2:** Apply Gaussian filter to the HSV image in order to remove the noise.

**Step3:** To detect objects having a selected color *Dcol*, threshold the Gaussian filtered image using equation (1):

$$I_{Step3} = \begin{cases} 1 & Dcol_{\min} \leq I_{Step2} \leq Dcol_{\max} \\ 0 & Otherwise \end{cases} \tag{1}$$

Where $I_{Step2}$ and $I_{Step3}$ are the images obtained after applying Step2 and Step3 respectively. $Dcol_{min}$ and $Dcol_{max}$ are the minimum and maximum values in the range of the chosen color to be detected, *Dcol*. The values of this range for the detected colors in the developed application are shown in Table 1, these values are determined after many experiments on some tested images.

**Step4:** Apply Erosion and Dilation for smoothing the boundary.

**Step5:** Find contours and draw them with a color value from the *Dcol* range to highlight the detected objects.

Figure 1, shows the output of the steps of the proposed algorithm to detect Green color objects in an image.

### 3.2. OBJECT DETECTION BASED ON ITS SHAPE

The application is capable of detecting circles, rectangles, triangles, and squares. Any shape other than these shapes is detected as "Other". The proposed algorithm for object detection based on its shape is as following:

**Step1:** The RGB color image that is loaded from the mobile gallery is converted to gray scale image.

**Step2:** Apply Gaussian filter to smooth the image in order to remove the noise.

**Step3:** Apply Canny edge detection algorithm for better accuracy to the Gaussian filtered image.

Table 1. The detected HSV colors range

| Color | Minimum Value | | | Maximum Value | | |
|---|---|---|---|---|---|---|
| | **H** | **S** | **V** | **H** | **S** | **V** |
| **Black** | 0 | 0 | 0 | 0 | 0 | 0 |
| **White** | 0 | 0 | 236 | 0 | 0 | 255 |
| **Gray** | 0 | 0 | 1 | 255 | 50 | 235 |
| **Blue** | 0 | 50 | 0 | 41 | 255 | 255 |
| **Green** | 42 | 0 | 0 | 88 | 255 | 255 |
| **Yellow** | 89 | 50 | 0 | 97 | 255 | 255 |
| **Orange** | 98 | 50 | 0 | 119 | 255 | 210 |
| **Beige** | 99 | 0 | 0 | 120 | 11 | 255 |
| **Red** | 120 | 140 | 0 | 128 | 255 | 255 |
| **Pink** | 120 | 0 | 241 | 149 | 255 | 255 |
| **Violet** | 150 | 0 | 0 | 200 | 255 | 255 |

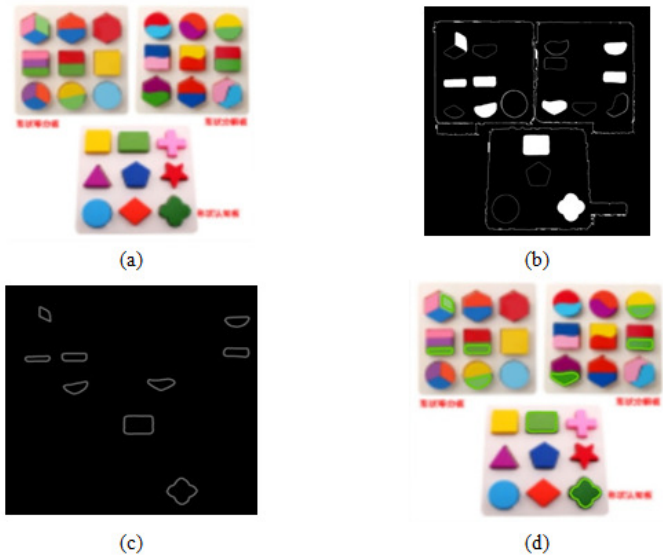

(a)        (b)

(c)        (d)

Figure 1. Green color objects detection
(a) RGB input image  (b) Thresholded image (c) Contour finding after erosion and dilation  (d) The green detected objects highlighted by a green contour

**Step4:** To detect circles, apply the Circle Hough Transform [11][12], and draw them.

**Step5:** To detect triangles, rectangles, squares, or Others:

1. Find contours and apply Douglas-Peucker algorithm [13] to approximate a contour shape to another shape with less number of vertices depending on a specified precision.
2. Check the approximated vertices value:
   - If it is three, then the shape is a triangle.
   - If it is four, the shape is a square or a rectangle or any polygonal have Four vertices. To specify the square and the rectangle, calculate the angle

> between two vertices by the dot product. Then to specify which one is a square and which one is a rectangle, calculate the ratio between the width and the height. Any other polygonal have four vertices is detected as "Other".

- • If it is greater than four, the shape is detected as "Other"
3. Draw the contours of the detected shapes.

### 3.3. OBJECT DETECTION BASED ON ITS LOCAL FEATURES

The proposed algorithm for object detection based on its BRISK local features consists of the following steps:

**Step1:** The object and scene RGB color images that are loaded from the mobile gallery are converted to gray scale images.

**Step2:** Extract the keypoints by BRISK detector [14] from both object and scene images.

**Step3:** Compute the descriptors for all keypoints by BRISK descriptor [14] for both object and scene images.

**Step4:** Match descriptors in the object image to the similar one in the scene image. This is done using hamming distance [14].

**Step5:** Select good matches by adjusting the minimum distances between keypoints, then find the number of good matches. A value of 50 for the number of good matches is set in the proposed algorithm. This value is determined based on experiments on some tested images.

**Step6:** To draw a polygon around the detected object, Find Homography [15] to find the transform between the matched keypoints and apply Perspective Transform [16] to map the points.

Figure 2, shows the output of the steps of the proposed algorithm to detect an object image in a scene image using BRISK local features.

## 4. EXPERIMENTAL RESULTS

For developing the proposed Android application, Eclipse IDE and OpenCV3 Library software tools are used. Then the application is ported and tested on Galaxy S3, S6, and Note1 Smartphones. Figure 3 shows the application icon and Main menu screen interface, the user can choose to detect an object by color feature, by shape feature, or by local features.

When "By Color Feature" is selected, the interface for Color detection will appear, as shown in Figure 4. A picture from the mobile gallery can be loaded then colors to be detected can be selected from a color list. Finally, the detected colors are highlighted in the loaded image. When "By Shape Feature" is selected, the interface, shown in Figure 5, will appear. A picture from the mobile gallery can be loaded then shapes to be detected can be selected from a shape list. Finally, the detected shapes are highlighted in the loaded image. When "By Local Features" is selected, the interface, shown in Figure 6, will appear. The object picture and the scene picture are loaded from the mobile gallery then the matched BRISK keypoints between the two pictures are highlighted by connected green lines, and the detected object is surrounded by a green polygon in the scene image.

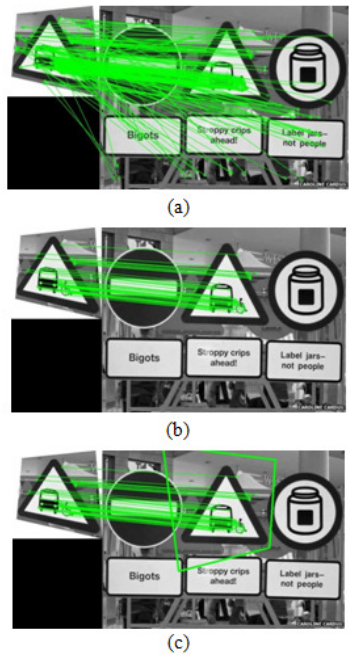

Figure 2. Objects detection by BRISK local features
(a) All matched keypoints in object and scene images (b) Minimum distance Matches,50 in our algorithm
(c) Draw polygon around the detected object in the scene image

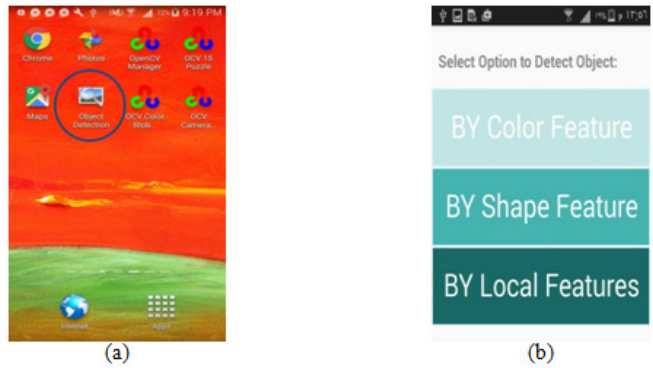

Figure 3. The Application interface
(a) Icon  (b) Main menu

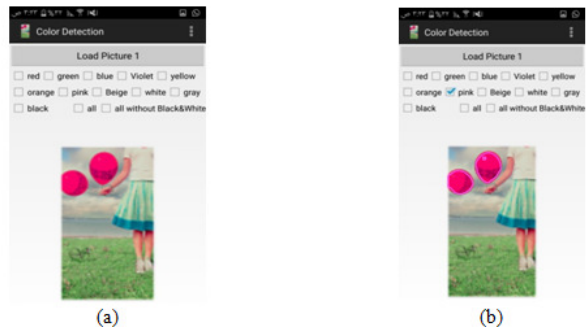

Figure 4. Color detection interface
(a) Load picture  (b) The selected "Pink" color is detected and highlighted in the loaded image

The application is tested using many different images. Figure 7, figure 8, and figure 9, show the output of the object detection by color, shape, and BRISK features respectively, on some sample images.

For color detection, the tested images are divided into two sets: uniform background images as figure 7 (a), (b), and (c) and complex background images as figure 7(e). All colors listed in Table 1 are correctly detected in the tested images.

For shape detection, the tested images are divided into three sets: Standard geometric shapes on a uniform background images as figure 8 (a), (b), (c), and (d), Compound and overlapped shapes on a uniform background images as figure 8(e), and (f), and standard shapes on a complex background images as figure 8(g). The results in figure 8 show that the shape detection algorithm correctly detected shapes on images that have standard geometric shapes on a uniform background, but it fails to detect some shapes in images with compound and overlapped shapes on a uniform background, and in images with standard shapes on a complex background. This is occurs for shapes that do not have complete connected edges or there are overlaps between them. Extra image preprocessing is required to resolve these issues.

For object detection by BRISK local features, the tested images are divided into four sets: Scaled object images as Figure 6(a), and Figure 9(a), Rotated object images as figure 2(c), Different view point object images as Figure 9(b), and Face images as Figure 9(c). The developed application correctly detected objects in images with different transformations like scaling and rotation, with different view point, and with faces. Table 2 summarizes the experimental testing results.

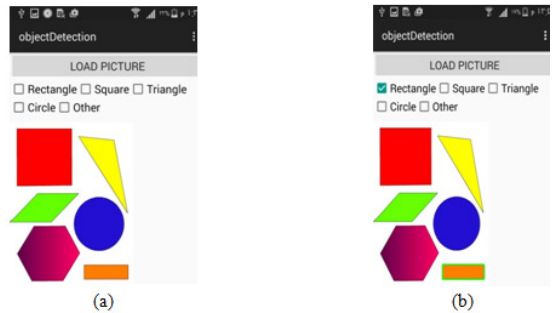

Figure 5. Shape detection interface
(a) Load picture  (b) The selected "Rectangle" shapes are detected and highlighted in the loaded image

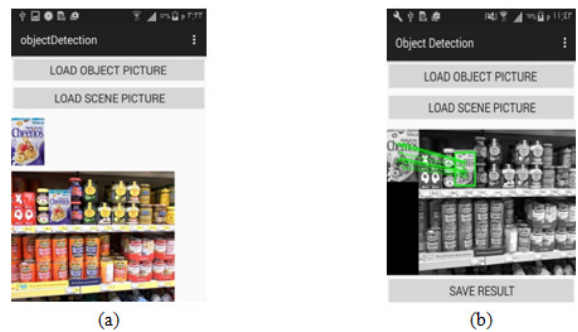

Figure 6. Object detection by BRISK local features interface
(a) Load object and scene pictures  (b) The matched BRISK keypoints are connected by green lines, and the detected object is surrounded by a green polygon in the scene image

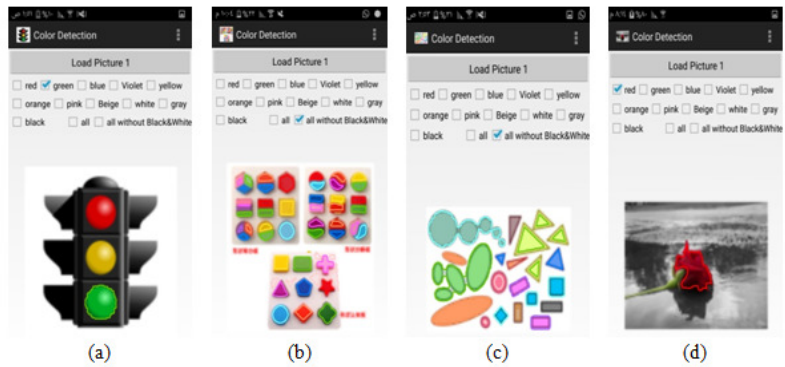

Figure 7. Color detection output on some sample images

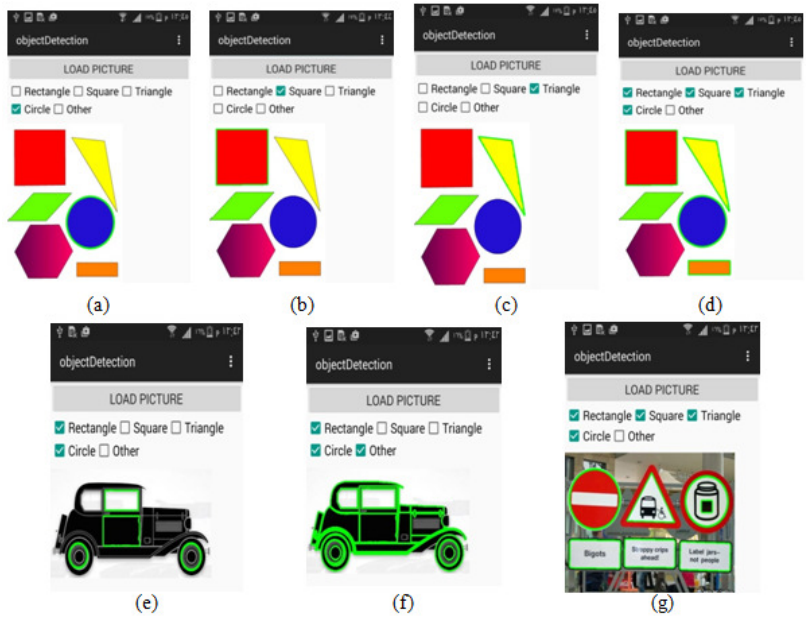

Figure 8. Shape detection output on some sample images

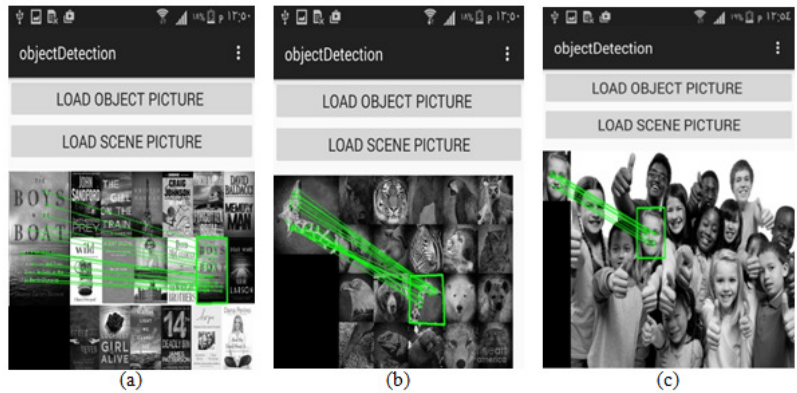

Figure 9. Object detection by BRISK output on some sample images

Table 2. Experimental Testing Results Summery.

| Object detection by: | Tested Images | Results |
|---|---|---|
| **Color features** | Uniform background images as fig.7 (a), (b), and (c) | Selected colors are correctly detected |
| | Complex background images as fig.7 (e) | |
| **Shape features** | Standard geometric shapes on uniform background images as fig.8 (a), (b), (c), and (d) | Selected standard shapes (Circle, Triangle, Rectangle, and Square) are correctly detected |
| | Compound and overlapped shapes on uniform background images as fig.8 (e), and (f) | Some Circles and Rectangles in fig.8.(e) are not detected, they are detected as "others" in fig.8(f) |
| | standard shapes on complex background images as fig.8 (g) | Some shapes are not detected |
| **BRISK local features** | Scaled object images as Fig.6 (a), Fig.9 (a) | Object image is correctly detected in the scene image |
| | Rotated object images as fig.2 (c) | |
| | Different view point object images as Fig.9 (b) | |
| | Face images as Fig.9 (c) | |

## 5. CONCLUSIONS

In this paper an Android application that is able to detect objects in an image that is loaded from the mobile gallery, based on its color, shape, or local features was developed. For color detection the image is processed in the HSV color domain for better color detection. Circular shapes are detected using Circular Hough Transform and other shapes are detected using Douglas-Peucker algorithm. BRISK local features are used to match an object image in a scene image. The application is ported and tested on Galaxy S3, S6, and Note1 Smartphones. Based on the experimental results, the application is capable of detecting correctly eleven different colors, detecting two dimensional geometrical shapes including circles, Rectangles, Triangles, and squares, and correctly match local features of object and scene images for different conditions.

As a future improvements, the shape detection algorithm could be modified to detect partial and overlapped shapes, and the application could be used as a part of e-learning application.

## AUTHORS

**Lamiaa A. Elrefaei** received her B.Sc. degree with honors in Electrical Engineering (Electronics & Telecommunications) in 1997. The M.Sc. in 2003 and Ph.D. in 2008 both in Electrical Engineering (Electronics). All from Faculty of Engineering at Shoubra, Benha University, Egypt. She has held a number of faculty positions at Benha University, as Teaching Assistant from 1998 to 2003, as an Assistant Lecturer from 2003 to 2008, and as Lecturer (equivalent to Assistant Professor) from 2008 to date. She is currently serving as an Assistant Professor at King Abdulaziz University, Jeddah, Saudi Arabia. Her research interests include Computational Intelligence, Image Processing, Computer Vision, and Pattern Recognition.

**Mona Omar Al-musawa** received her B.Sc. degree in computer science from Sana'a University, Yemen in 2010, she is currently M.Sc. student at Faculty of Computing and Information Technology, King Abdulaziz University, Saudi Arabia.  Her research interests are computer vision, machine learning and virtual reality.

**Norah Abdullah Al-gohany** received her bachelor degree with first honors in computer science in 2014 from Faculty of Computer Science and Engineering in Medina, Taibah University, Saudi Arabia. She is currently a Master student at King Abdulaziz University, Jeddah, Saudi Arabia. Her research interests include Programming languages, artificial intelligence, encryption and image processing.